%% file: main.tex
\documentclass[10pt,twocolumn,letterpaper]{article}

\usepackage{times}
\usepackage{epsfig}
\usepackage{graphicx}
\usepackage{amsmath}
\usepackage{amssymb}
\usepackage{booktabs}
\usepackage{indentfirst}
\usepackage{subfig}

\usepackage[pagenumbers]{cvpr}      % To produce the REVIEW version
\input{preamble}

\definecolor{cvprblue}{rgb}{0.21,0.49,0.74}
\usepackage[pagebackref,breaklinks,colorlinks,citecolor=cvprblue]{hyperref}

\def\paperID{19} % *** Enter the Paper ID here
\def\confName{CVPR}
\def\confYear{2024}

\title{Image-based Detection of Segment Misalignment \\in Multi-mirror Satellites using Transfer Learning}

\author{C. Tanner Fredieu$^1$$^,$$^2$, Jonathan Tesch$^2$, Andrew Kee$^2$, David Redding$^2$\\
$^1$Virginia Tech, Blacksburg, Virginia\\
$^2$NASA Jet Propulsion Laboratory, California Institute of Technology, Pasadena, California\\
{\tt\small Email: christianf@vt.edu, jonathan.a.tesch@jpl.nasa.gov,
}\\{\tt\small andrew.g.kee@jpl.nasa.gov, david.c.redding@jpl.nasa.gov
}
}

\begin{document}
% \maketitle
% \input{sec/0_abstract}    
% \input{sec/1_intro}
% \input{sec/2_formatting}
% \input{sec/3_finalcopy}
% {
%     \small
%     \bibliographystyle{ieeenat_fullname}
%     \bibliography{main}
% }


% \end{document}

\maketitle
\thispagestyle{empty}

%%%%%%%%% ABSTRACT
\begin{abstract}
   In this paper, we introduce a system based on transfer learning for detecting segment misalignment in multimirror satellites, such as future CubeSat designs and the James Webb Space Telescope (JWST), using image-based methods. When a mirror segment becomes misaligned due to various environmental factors, such as space debris, the images can become distorted with a shifted copy of itself called a "ghost image". To detect whether segments are misaligned, we use pre-trained, large-scale image models trained on the Fast Fourier Transform (FFT) of patches of satellite images in grayscale. Multi-mirror designs can use any arbitrary number of mirrors. For our purposes, the tests were performed on simulated CubeSats with 4, 6, and 8 segments. For system design, we took this into account when we want to know when a satellite has a misaligned segment and how many segments are misaligned. The intensity of the ghost image is directly proportional to the number of segments misaligned. Models trained for intensity classification attempted to classify N-1 segments. Across eight classes, binary models were able to achieve a classification accuracy of 98.75\%, and models for intensity classification were able to achieve an accuracy of 98.05\%.
\end{abstract}

%%%%%%%%% BODY TEXT
\section{Introduction}

On 25 December 2021, the James Webb Space Telescope (JWST) was launched into space by the United States National Aeronautics and Space Administration (NASA). One of the key features of JWST was its 18-hexagonal mirror design to act as one 6.5 meter mirror. The use of smaller, segmented mirrors to act as one larger mirror has been in use in ground-based telescopes, such as the Keck and Hobby-Eberly telescopes, but JWST is the first space-based telescope to use this configuration. Due to its resounding success, segmented mirrors are being investigated for use in other different types of satellites such as CubeSats for Low-Earth orbit (LEO) operations. This type of satellite is seen as affordable and low-cost options with a wide range of applications, such as image classification, \cite{9553108}, \cite{9239508}, \cite{9884827}. With this being a recent development in space technologies, research and development is new in the area. One of the primary concerns of such satellites is how to maintenance them once deployed in orbit. For those that reside in space beyond Earth's orbit, such as JWST, maintenance is currently impossible due to the vast distances and costs of any such maintenance-based missions. This is why development of the JWST was prolonged as all the instruments had to be guaranteed to be completely operational once launched. For satellites such as CubeSats in orbit, maintenance could be possible in the future but is difficult in the immediate without the need for additional launches. For now, any maintenance of the satellite must be able to be performed by the satellite's systems either remotely or autonomously.

The ability to provide this on-board analysis of image systems is the primary motivation behind this work, and our contributions are summarized below:
\begin{itemize}
    \item We examine the use of transfer learning and pre-trained models to perform anomaly detection using only ground images from satellites to check mirror alignment.
    \item We provide a solution for detection of ghost components in images that is not hardware-based but software and image-based.
    \item To our knowledge, we are the first to use a novel method using deep learning and satellite ground images to monitor satellite/telescope operational status.
\end{itemize}

% \break

\section{Related Work}
Aforementioned, work directly related to the application of computer vision models to these types of multi-mirror satellites is limited since the only active one in service from any space agency is the JWST operated by NASA. However, much work has been done in the field of image classification, blur detection, and detection of irregularities in images \cite{4663021}, \cite{7894491}, \cite{9259898}. This work and others are what we used as the basis for system development later on. In this section, methods using both machine learning and non-machine learning techniques along with the related work that uses these techniques will be discussed.

\subsection{Non-machine Learning Methods}
The two most notable and popular methods of computer vision for the detection of irregularities, especially for the detection of blur in grayscale images, are to take a Fast Fourier Transform (FFT) of an image and to use the Laplacian kernel.

Both methods were initially tried individually in our experimentation without success before deep learning techniques were implemented. The primary reason for the failure of these methods was finding a suitable threshold value to detect low-intensity ghosting in the images.

\subsubsection{Laplacian Kernel Method}
The first method discussed is using the Laplacian kernel which is a 3x3 matrix shown below along with the derivation of the discrete Laplacian using finite approximation:

\begin{equation}
    \nabla^2 = \frac{\partial}{\partial x^2}+\frac{\partial}{\partial y^2}
\end{equation}

\begin{equation}
    f'(x) = \lim_{h\to 0} \frac{f(x+h)-f(x)}{h}
\end{equation}
    
\begin{equation}
    \frac{\partial}{\partial x^2} = f(x+1) - 2f(x) + f(x-1)
\end{equation}
    
\begin{equation}
    \frac{\partial}{\partial y^2} = f(y+1) - 2f(y) + f(y-1)
\end{equation}

\begin{equation}
\begin{pmatrix}
0 & 1 & 0  \\
1 & -4 & 1  \\
0 & 1 & 0  
\end{pmatrix}
=\begin{pmatrix}
0 & 0 & 0  \\
1 & -2 & 1  \\
0 & 0 & 0  
\end{pmatrix} +
\begin{pmatrix}
0 & 1 & 0  \\
0 & -2 & 0  \\
0 & 1 & 0  
\end{pmatrix}
\end{equation}

% \begin{figure}[h]
% \begin{center}
% \begin{tabular}{cc}
% $\begin{pmatrix}
% 0 & 1 & 0  \\
% 1 & -4 & 1  \\
% 0 & 1 & 0  \nonumber
% \end{pmatrix} $ &
% % \includegraphics[scale = 0.3, keepaspectratio = 0.3]{variance} \\
% \end{tabular}
% \end{center}
% \end{figure}

The variance of the image after the Laplacian kernel was applied in \cite{7894491} and \cite{7397496} to assess image quality and applied in \cite{9952148} and \cite{9166205} to detect blur. Determining if an image is blurry using this method depends on the variance outputted. If a high variance is detected from the image, then the image quality is high. If the variance is low, the quality is low, suggesting possible problems with the camera or other hardware. The benefit of this method is that it is fast and efficient. The drawback is the need to find a threshold value as well as not being as effective as other more computationally costly methods such as FFT.

\subsubsection{Fast Fourier Transform Method}
The second method discussed uses the FFT of an image to determine whether irregularities have occurred. By taking the FFT of an image along with its magnitude, the distribution of frequencies can be found.

\begin{equation}
F(m,n) = \sum_{m=0}^{M-1} \sum_{n=0}^{N-1} f(x,y)e^{-j2\pi(\frac{um}{M}+\frac{vn}{N}})
\end{equation}

This method has the advantage of being able to accurately detect a wider range of variance in the magnitude of the image compared to other methods such as using a Laplacian kernel. The use of FFT has been used successfully in image quality assessment applications \cite{4663021}, \cite{7025113}. However, it suffers from the same initial problem as the Laplacian method, where a threshold is needed to be manually calculated and it can miss small image distortions such as the ghost component. An example of the FFT performed on an image from the dataset used is illustrated in Figure 1.

\begin{figure}
\begin{center}
% \fbox{\rule{0pt}{2in} \rule{.9\linewidth}{0pt}}
    \subfloat[\centering Satellite image]{{\includegraphics[scale=0.035]{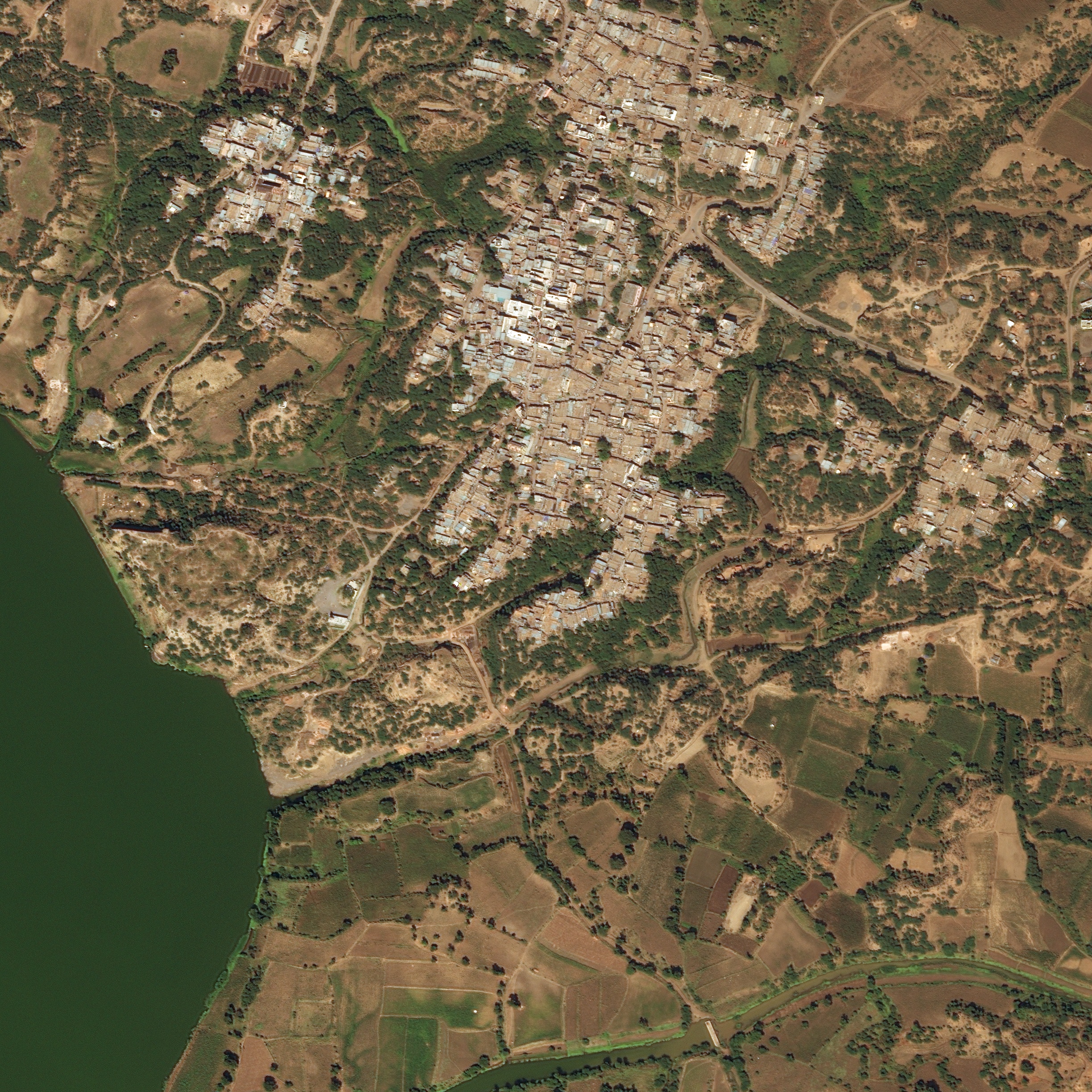} }}%
    \qquad
    \subfloat[\centering FFT of image]{{\includegraphics[scale=0.035]{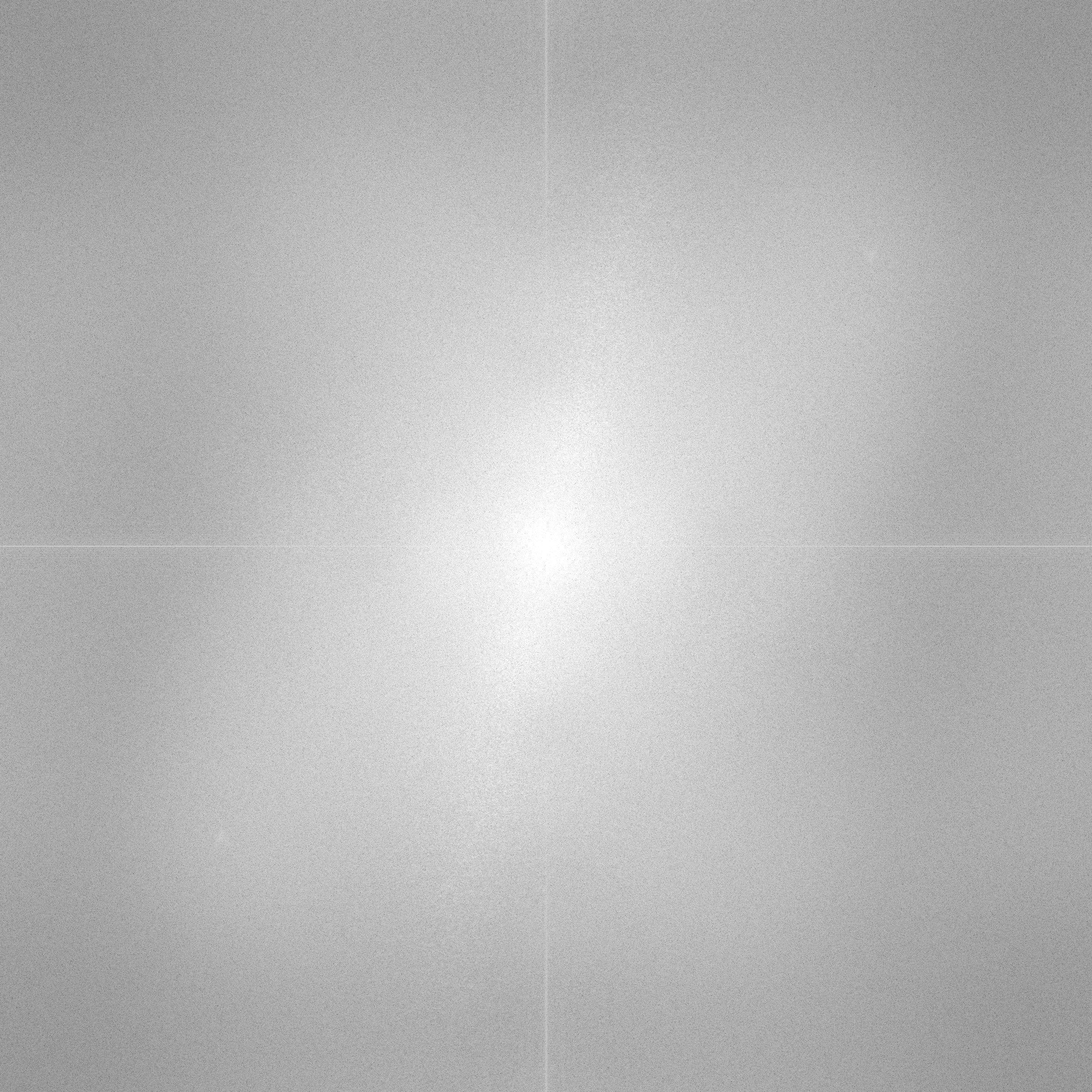} }}%
\end{center}
   \caption{Satellite image before and after FFT}
% \label{fig:short}
\end{figure}

\subsection{Machine Learning Methods}
 Deep learning has proven effective for a wide range of applications in computer vision, language processing, generative models, etc. These techniques have also been widely utilized in the fields of astronomy, space and planetary exploration, and satellite imaging. Such applications range from target tracking \cite{9836753}, weather monitoring \cite{9696140}, \cite{10002657}, satellite communications \cite{9622204}, \cite{9449160}, imaging of Earth's and other planetary surfaces \cite{9844568}, and analysis of deep space data such as solar data \cite{9812657}.

To address the problem of determining an optimal threshold manually, as with Laplacian and FFT methods to distinguish blurred images from non-blurred ones, deep learning methods have been used to great effect. This can be achieved in a few ways. The simplest way is to use a deep neural network by itself as in \cite{9308594}, \cite{9914850}, \cite{8261503}, and \cite{8451765}. More complex and effective methods involve combining deep neural networks with others, such as Laplacian and FFT methods, to increase performance. \cite{10206110}, \cite{resnset} and \cite{10306622} give examples of convolutional neural networks (CNN) paired with Laplacian kernels used to provide enhanced features in preprocessing, while \cite{10157161} gives an example of the FFT method paired with a CNN. The primarily benefit of these approaches compared to the more simpler one is the higher degree of unique feature extraction that can occur using these methods in the preprocessing steps to improve model performance.

\begin{figure*}
\begin{center}
% \fbox{\rule{0pt}{2in} \rule{.9\linewidth}{0pt}}
\includegraphics[scale=0.45]{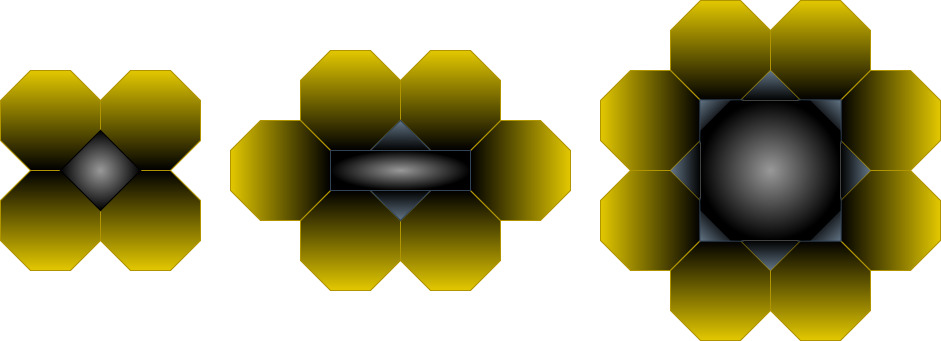}
\end{center}
   \caption{Simulated arbitrary mirror configurations for 4, 6, and 8 mirrors}
\label{fig:short}
\end{figure*}

For our purposes, we will also focus on the paradigm of transfer learning. Transfer learning looks to follow the way humans can transfer experience and knowledge from one task to another to adapt and learn more efficiently \cite{6847217}. With the need for large-scale models for many modern solutions to vision tasks, the use of transfer learning with pre-trained models for cross-domain learning is prevalent in many areas of research. It has been widely applied, as of recently, in the aforementioned fields \cite{10074554}, \cite{9836753}, \cite{9259898}, \cite{9696140}, \cite{10054990}.

While all these methods show the success of deep learning and transfer learning in various applications involving satellite systems, they are all focused on object status and/or information present in the image. There does not seem to be work in the area of determining the system and hardware status or quality from the images alone. As we mentioned previously, to our knowledge, this work is the first to focus on using the deep learning with images to assess the system itself instead of the image.
%-------------------------------------------------------------------------
% \begin{figure*}
% \begin{center}
% % \fbox{\rule{0pt}{2in} \rule{.9\linewidth}{0pt}}
% \includegraphics[scale=0.5]{Images/JPL-Paper-Page-5.jpg}
% \end{center}
%    \caption{Simulated arbitrary mirror configurations for 4, 6, and 8 mirrors}
% \label{fig:short}
% \end{figure*}

\section{Implementation}
\subsection{Experimental Setup}
As stated previously, the detection of ghost components in images does not have the depth of literature as other tasks in the fields of computer vision and image processing. So to create a baseline performance for our model, we added Gaussian blur to the images, the specifics will be detailed later, in the first experimentation then tested our model against other models used for blur detection. This initial experimentation focused on binary classification to examine if the model was able to determine whether an image contained blur or not before moving on to the primary experimentation.

For implementation of our model in the blur detection experimentation, we focused on three different approaches for our preprocessing of the images. First, we implemented a straightforward method to input the 2448x2448 images directly into the model. No preprocessing was performed on the images to test the effect this had on performance. Second, we took the first approach and apply FFT to all the images then take the magnitude of each before sending them to the model. Third, each image from the second approach was cut into patches of individual 266x266 images before the FFT was applied and then sent to the model. This method was inspired by \cite{8451765}. 

For the primary experimentation, we took the best performing approach from the initial experimentation then applied it to this one. The only difference is the use of ghosting components added to the images instead of a Gaussian blur kernel.
% the second approach and apply FFT to all the patches then take the magnitude of each. The initial experimentation focused on binary classification of attempting to detect whether any misalignment has occurred or not. 
This part of the experimentation also attempted to expand the classification ability by classifying whether a misalignment has occurred, and if so, how many mirrors are misaligned from a simulated segmented mirror instead of only the binary classification from the first experimentation.

For our CubeSat design, we simulated three different mirror configurations outlined in Figure 2. These configurations are purely arbitrary, but they are meant to illustrate how a multimirror design can be presented on a satellite. 
\subsection{Datasets}
Since there is no common, benchmark dataset for images with ghost components, our team took a satellite dataset from Kaggle and added the desired effects for our purposes. The dataset chosen was the DeepGlobe Land Cover Classification Dataset \cite{DeepGlobe18}. Due to the images in the DeepGlobe dataset containing three channels of RGB, the entire dataset is first converted to grayscale.

For the initial experimentation with blur detection, Gaussian blur with sigma values from the set {1,2,3,4,5} was added to the grayscale images of the dataset. For the primary experimentation, we added ghosting to each image using the following preprocessing steps. 
\begin{enumerate}
    \item A copy of the image is first made
    \item A translation matrix, below, containing $t_x$ and $t_y$ is used to provide a pixel offset on the copy. $t_x$, $t_y$ $\in$ [0,15]. In the first experiment, the values of $t_x$ and $t_y$ are randomized while, in the second, the values are $\propto$ I.

        \begin{equation}
            \bf{T} = \begin{pmatrix}
            1 & 0 & t_x  \\
            0 & 1 & t_y  \\
            0 & 0 & 1  
            \end{pmatrix} 
        \end{equation}
    \item Mirror offset in I $\in$ [0, $1-\frac{1}{N}$] , where N is the number of mirror segments on the satellite.
    \item The copy is then added to the original with the intensity of the original being (1-I), and the intensity of the copy being I.
\end{enumerate}

An example of an image with the ghost component added is illustrated in Figure 3.

\begin{figure}[t]
\begin{center}
% \fbox{\rule{0pt}{2in} \rule{.9\linewidth}{0pt}}
    \subfloat[\centering Before ghosting]{{\includegraphics[scale=0.35]{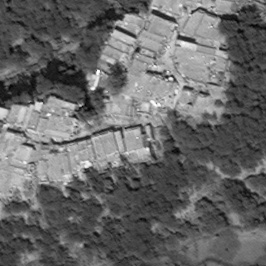} }}%
    \qquad
    \subfloat[\centering After ghosting]{{\includegraphics[scale=0.35]{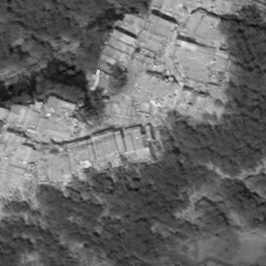} }}%
\end{center}
   \caption{Image before and after ghosting}
% \label{fig:short}
\end{figure}

\begin{figure*}
\begin{center}
% \fbox{\rule{0pt}{2in} \rule{.9\linewidth}{0pt}}
\includegraphics[scale=0.488]{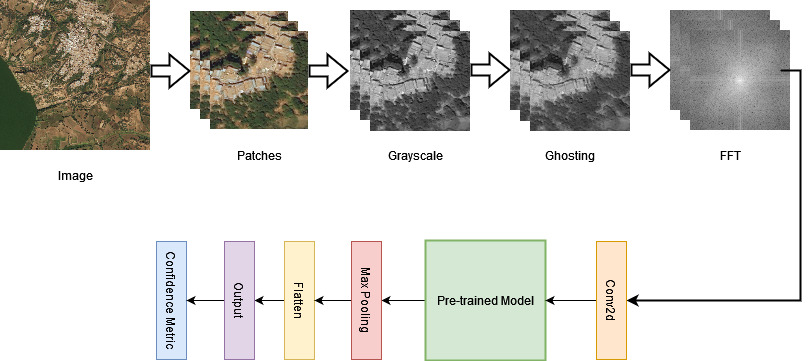}
\end{center}
   \caption{Final system design}
% \label{fig:short}
\end{figure*}

% \break

% \begin{figure}[t]
% \begin{center}
% % \fbox{\rule{0pt}{2in} \rule{.9\linewidth}{0pt}}
%     \subfloat[\centering Before ghosting]{{\includegraphics[scale=0.35]{Images/wacv non-ghost image.png} }}%
%     \qquad
%     \subfloat[\centering After ghosting]{{\includegraphics[scale=0.35]{Images/wacv ghost image.png} }}%
% \end{center}
%    \caption{Image before and after ghosting}
% % \label{fig:short}
% \end{figure}

\subsection{Model Architecture and System Design}
For our model architecture, we make use of pre-trained models as our primary component. We chose three models because of their availability and commonality. These models are VGG-19, ResNet-50, and ResNet-101. VGG-19 was chosen specifically to test a system that. ResNet-50 was chosen to test the effect of residual connections on the performance while ResNet-101 was to test if deepening the network had any effect.

Since all of the pre-trained models were trained on colored images contained 3 channels of RGB, a 2D convolutional layer with a filter size of 3 and a kernel size of 3 was used to transform the 1 channel grayscale images into 3 by simply copying the channel three times. The images are then fed to the pre-trained model with max pooling and flatten layers before being sent to the output layer. Max pooling was chosen as we are interested in the max differences between the pixels in the images. The general and final system architecture is illustrated in Figure 4.  

% \begin{figure*}
% \begin{center}
% % \fbox{\rule{0pt}{2in} \rule{.9\linewidth}{0pt}}
% \includegraphics[scale=0.45]{Images/JPL-Paper-Page-5.jpg}
% \end{center}
%    \caption{Simulated arbitrary mirror configurations for 4, 6, and 8 mirrors}
% \label{fig:short}
% \end{figure*}

% \begin{figure}[t]
% \begin{center}
% % \fbox{\rule{0pt}{2in} \rule{.9\linewidth}{0pt}}
%     \subfloat[\centering Satellite patch image]{{\includegraphics[scale=0.35]{Images/wacv non-ghost image.png} }}%
%     \qquad
%     \subfloat[\centering FFT of patch]{{\includegraphics[scale=0.35]{Images/wacv ghost image.png} }}%
% \end{center}
%    \caption{Example of a short caption, which should be centered.}
% % \label{fig:short}
% \end{figure}

% \begin{figure}[t]
% \begin{center}
% % \fbox{\rule{0pt}{2in} \rule{.9\linewidth}{0pt}}
%     \subfloat[\centering Satellite patch image]{{\includegraphics[scale=0.35]{Images/wacv pre-FFT patch image.png} }}%
%     \qquad
%     \subfloat[\centering FFT of patch]{{\includegraphics[scale=0.35]{Images/wacv FFT patch image.png} }}%
% \end{center}
%    \caption{Example of a short caption, which should be centered.}
% % \label{fig:short}
% \end{figure}

% \begin{figure}[t]
% \begin{center}
% % \fbox{\rule{0pt}{2in} \rule{.9\linewidth}{0pt}}
%     \subfloat[\centering Satellite patch image]{{\includegraphics[scale=0.35]{Images/wacv non-ghost image.png} }}%
%     \qquad
%     \subfloat[\centering FFT of patch]{{\includegraphics[scale=0.35]{Images/wacv ghost image.png} }}%
% \end{center}
%    \caption{Example of a short caption, which should be centered.}
% % \label{fig:short}
% \end{figure}

\begin{figure}[t]
\begin{center}
% \fbox{\rule{0pt}{2in} \rule{.9\linewidth}{0pt}}
\includegraphics[scale=0.35]{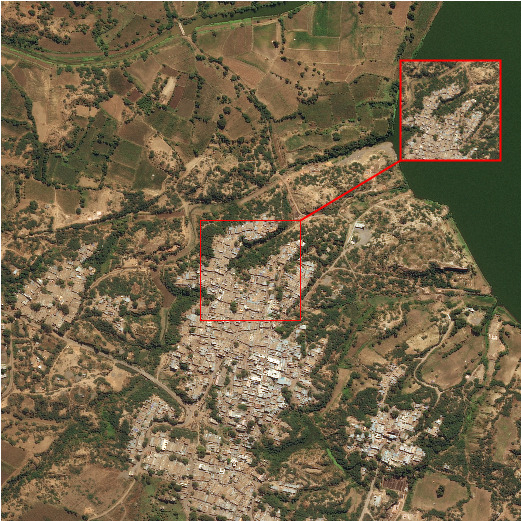}
\end{center}
   \caption{Patch image being taken from satellite image}
% \label{fig:short}
\end{figure}

% \begin{figure*}
% \begin{center}
% % \fbox{\rule{0pt}{2in} \rule{.9\linewidth}{0pt}}
% \includegraphics[scale=0.45]{Images/JPL-Paper-Page-7.jpg}
% \end{center}
%    \caption{Simulated mirror configurations for 4, 6, and 8 mirrors}
% % \label{fig:short}
% \end{figure*}

%-------------------------------------------------------------------------
\section{Experimental Results}

For the first part of our experimentation, we test the three different approaches outlined in the Implementation section with the DeepGlobe dataset with Gaussian blur.
% we focused on a binary detection system of recognizing whether a misalignment had occurred. 
Table 1 illustrates the performance of the three models for each approach. The initial results of the first approach show less than optimal performance for the three models. The most optimal of them was only able to achieve slightly above the accuracy of 70\%. Incorporation of FFT in image preprocessing gave an increase in performance of 6\% on average, but overall accuracy was still lower than expected. The third approach came about after investigating the FFT of the entire ground image. Once the FFT was applied, we observed little variance in magnitude, as illustrated in Figure 1.b. To resolve this, we took patches of each image which is illustrated in Figure 5. The difference in magnitude variance using this method is illustrated in Figure 6.b. Once this extra preprocessing step was added, the model performance increased dramatically. As illustrated in Table 1, accuracy across all models increased above 95\% with the optimal model able to perform with accuracy over 98\% from a previous performance of 76.67\%

\begin{table}[t]
\caption{Approach comparisons}
  \begin{center}
    {\small{
\begin{tabular}{lllr}
\toprule
Model & Entire Image  & FFT/Entire  & FFT/Patches \\
\midrule
VGG-19 & 0.6677 &  0.7264 & 0.9574\\
ResNet-50 & 0.6852 & 0.7477 & 0.9878\\
ResNet-101 & \bf{0.7089} & \bf{0.7667} & \bf{0.9889}\\
\bottomrule
\end{tabular}
}}
\end{center}

\end{table}

\begin{table}[t]
\caption{Model comparison on image patches}
  \begin{center}
    {\small{
\begin{tabular}{lllr}
\toprule
Model & Accuracy   \\
\midrule
CNN \cite{9308594} & 0.6776 \\
ResNet-50 \cite{resnset} & 0.9583 \\
ResNet-50 (ours) & 0.9878 \\
ResNet-101 (ours) & \bf{0.9889} \\
\bottomrule
\end{tabular}
}}
\end{center}

\end{table}

Of the three models, the model based on the ResNet-101 architecture was the most optimal when it came to performance in all categories having an average 4\% increase over the VGG-19 model and 2\% increase over the ResNet-50 in the first and second approaches. However, the model based on the ResNet-50 architecture showed only slight performance degradation compared to the third and final approach showing that a deeper network gives only small improvements in performance.

% The model is able to detect the ghost component in a majority of the images, but it fails in detecting minimum amounts of misalignment when the intensity falls below 25\%. The highest accuracy we were able to achieve using this approach was 70.89\% in the four-mirror configuration using ResNet-101. ResNet-101 being the optimal model will be a reoccurring theme.

% \begin{table}[t]
% \caption{No FFT - patches }
%   \begin{center}
%     {\small{
% \begin{tabular}{lllr}
% \toprule
% Model & 4 Mirrors & 6 Mirrors & 8 Mirrors \\
% \midrule
% VGG-19 & 0.7264 & 0.7262 & 0.7053\\
% ResNet-50 & 0.7477 & 0.7472 & 0.7456\\
% ResNet-101 & \bf{0.7667} & \bf{0.7560} & \bf{0.7555}\\
% \bottomrule
% \end{tabular}
% }}
% \end{center}

% \end{table}

% \begin{table}
% \caption{FFT - patches}
%   \begin{center}
%     {\small{
% \begin{tabular}{lllr}
% \toprule
% Model & 4 Mirrors & 6 Mirrors & 8 Mirrors \\
% \midrule
% VGG-19 & 0.9574 & 0.9572 & 0.9463\\
% ResNet-50 & 0.9878 & 0.9873 & 0.9769\\
% ResNet-101 & \bf{0.9889} & \bf{0.9880} & \bf{0.9875}\\
% \bottomrule
% \end{tabular}
% }}
% \end{center}

% \end{table}

% \break

% Our second approach used the FFT method discussed earlier to find differences in magnitude across images. 

Table 2 illustrates the results from the performances of the two comparison models taken from \cite{9308594} and \cite{resnset} along with the ResNet-50 and ResNet-101 versions of our model. These two models were chosen due to the model in \cite{9308594} not using any Laplacian or FFT methods in its preprocessing, similar to our approach one in the first experimentation, and the model in \cite{resnset} also being based on the ResNet-50 architecture and using a Laplacian method in its preprocessing. This allowed for further comparisons of a model with a novel architecture using no extra preprocessing, and a comparison between two of the same models, one using a Laplacian kernel and another (ours) using FFT. As shown in Table 2, the CNN model from \cite{9308594} performed the worst among the models with a performance similar to the entire image approach in Table 1. The model from \cite{resnset} had similar performances to our models with the accuracy performance reaching 95.83\%. However, this is a 3\% lower performance when compared to our ResNet-50 and ResNet-101 models. Showing that the use of FFT in the preprocessing step provides more improvement than when using a Laplacian kernel.

% Table 2 illustrates our results. While there is an increase in accuracy of approximately 5\% across all models, maximum accuracy is still limited to 76.67\%. After investigating the FFT of the whole ground image, we observed little variance in magnitude as illustrated in Figure 1.b. To resolve this, we took patches of each image as previously discussed in the Methods section which is illustrated in Figure 5. The difference in magnitude variance using this method is illustrated in Figure 6.b. Once this extra preprocessing step was added, the model performance increased dramatically. As illustrated in Table 3, accuracy across all models increased above 94\% with the optimal model able to perform with accuracy over 98\%.

\begin{figure*}
\begin{center}
% \fbox{\rule{0pt}{2in} \rule{.9\linewidth}{0pt}}
    \subfloat[\centering Satellite patch image]{{\includegraphics[scale=0.5]{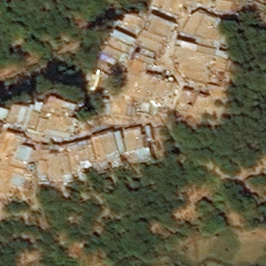} }}%
    \qquad
    \subfloat[\centering FFT of patch]{{\includegraphics[scale=0.5]{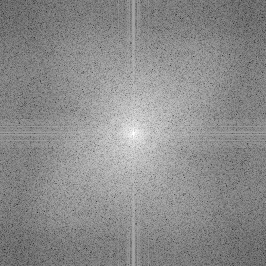} }}%
\end{center}
   \caption{FFT of image patch}
% \label{fig:short}
\end{figure*}

\begin{table}[t]
\caption{Intensity Classification - Random Offset}
  \begin{center}
    {\small{
\begin{tabular}{lllr}
\toprule
Model & 4 Mirrors & 6 Mirrors & 8 Mirrors \\
\midrule
VGG-19 & 0.5656 & 0.4396 & 0.3278\\
ResNet-50 & 0.6062 & 0.4534 & 0.3221\\
ResNet-101 & \bf{0.6231} & \bf{0.5124} & \bf{0.3567}\\
\bottomrule
\end{tabular}
}}
\end{center}

\end{table}

% \begin{table}[h]
% \caption{Intensity Classification with only misalignment classes}
%   \begin{center}
%     {\small{
% \begin{tabular}{lllr}
% \toprule
% Model & 4 Mirrors & 6 Mirrors & 8 Mirrors \\
% \midrule
% VGG-19 & 0.7285 & 0.7050 & 0.7024\\
% ResNet-50 & 0.7422 & 0.7284 & 0.7166\\
% ResNet-101 & \bf{0.7497} & \bf{0.7300} & \bf{0.7294}\\
% \bottomrule
% \end{tabular}
% }}
% \end{center}

% \end{table}

\begin{table}[t]
\caption{Intensity Classification - No Random Offset}
  \begin{center}
    {\small{
\begin{tabular}{lllr}
\toprule
Model & 4 Mirrors & 6 Mirrors & 8 Mirrors \\
\midrule
VGG-19 & 0.9534 & 0.9552 & 0.9415\\
ResNet-50 & 0.9827 & 0.9788 & 0.9754\\
ResNet-101 & \bf{0.9859} & \bf{0.9850} & \bf{0.9805}\\
\bottomrule
\end{tabular}
}}
\end{center}

\end{table}

The results of the primary experiment are illustrated in Tables 3 and 4. Our initial trials attempted to use the same selection method, randomization, for the values of $t_x$ and $t_y$ as for the binary models. However, this proved to be detrimental to performance as illustrated in Table 3. Even with the most optimal model, ResNet-101, accuracy performance was limited to 62.31\% at best and steadily decreased as more mirror segments are added with an accuracy performance of 35.67\% with 8 segments. 

After other test trials, we found that the reason for this performance drop was due to the selection method for the values of $t_x$ and $t_y$. Initially used to allow for a wide range of pixel offsets, the model was unable to distinguish class difference from the level of intensity of the ghost component alone. The successful trial set the values of $t_x$ and $t_y$ $\propto$ I which is beneficial as the value of the pixel offset being related to the intensity of the misalignment is more consistent with real-world affects. The results of that trial are illustrated in Table 4. As shown, the model was now more consistent with earlier results as accuracy performance reached 98.59\%. Using these patches of the images also had another added benefit. It allowed for the system to classify over multiple sections of an image providing a confidence metric for images to be assessed on whether a ghost component was present.

%-------------------------------------------------------------------------
\section{Future Work}
For our future work in this area, we plan to further investigate the possible environmental effects that can occur in ground images for better detection. We will also investigate different methods for preprocessing. Currently, we do all preprocessing manually by adding the ghost effects to images one by one. In the future, we will look at using a point-spread function to estimate the ghost component's intensity and offset. It would also be beneficial to evaluate the computation resources used by each model and to compare the trade-off in accuracy to efficiency in resource use.

% Second, we will investigate the relation between pixel offset and the number of mirror misaligned. For the initial experimentation, 

% Third, we will investigate into other architectures and/or modifications of our existing system.
%-------------------------------------------------------------------------
% \break
\section{Conclusions}
In conclusion, we have proposed a deep learning system based on pre-trained image models that is able to not only detect that a mirror misalignment has occurred, but also the intensity at which it has in multi-mirror satellites. By taking the FFT of patches of a ground image from satellites, we are able to achieve an accuracy of 98.75\% in detecting a general misalignment regardless of the mirrors affected and an accuracy of 98.05\% for the intensity of the misalignment. Also, we were able to use these patches to perform a confidence metric on the image to determine with a better degree of accuracy whether a ghost component was present in the image or not.
%-------------------------------------------------------------------------

% \section*{Acknowledgment}
% This research was carried out at the Jet Propulsion Laboratory, California Institute of Technology, under a contract with the National Aeronautics and Space Administration (80NM0018D004).
% \section{References}
{
    \small
    \bibliographystyle{ieeenat_fullname}
    \bibliography{main}
}
%-------------------------------------------------------------------------

\end{document}

%% file: preamble.tex
\usepackage[dvipsnames]{xcolor}